\documentclass[10pt,twocolumn,letterpaper]{article}

\usepackage[pagenumbers]{cvpr} 

\usepackage{graphicx}
\usepackage{amsmath}
\usepackage{amssymb}
\usepackage{booktabs}
\usepackage{multirow}

\usepackage{cuted}

\usepackage[pagebackref,breaklinks,colorlinks]{hyperref}

\newcommand{\boldstart}[1]{\noindent\textbf{#1}}
\newcommand{\boldstartspace}[1]{\vspace{0.1in}\noindent\textbf{#1}}

\usepackage[capitalize]{cleveref}
\crefname{section}{Sec.}{Secs.}
\Crefname{section}{Section}{Sections}
\Crefname{table}{Table}{Tables}
\crefname{table}{Tab.}{Tabs.}

\def\cvprPaperID{5064} 
\def\confName{CVPR}
\def\confYear{2022}

\begin{document}

\newcommand{\modelname}{{NeRFusion}} 

\title{\modelname: Fusing Radiance Fields for Large-Scale Scene Reconstruction}


\author{
Xiaoshuai Zhang$^{1\thanks{Research partially done during Xiaoshuai's internship at Adobe Research}}$
\,\,
Sai Bi$^2$
\,\,
Kalyan Sunkavalli$^2$ 
\,\,
Hao Su$^1$
\,\,
Zexiang Xu$^2$ \qquad \qquad \\
$^{1}$ University of California, San Diego \qquad $^{2}$ Adobe Research\\
{\tt\small {{\{xiz040,haosu\}}@eng.ucsd.edu \quad {\{sbi,sunkaval,zexu\}}@adobe.com \quad }
}
}

\newcommand{\B}[1]{\textbf{#1}}


\definecolor{salmon}{rgb}{1.0, 0.55, 0.41}
\newcommand{\jet}[1]{{\color{salmon}{Xiaoshuai: #1}}}
\newcommand{\zexiang}[1]{{\color{blue}{Zexiang: #1}}}
\newcommand{\sai}[1]{{\color{magenta}{Sai: #1}}}
\newcommand{\KS}[1]{{\color{green}{Kalyan: #1}}}

\maketitle

\begin{strip}\centering
\vspace{-6em}
\includegraphics[width=\textwidth]{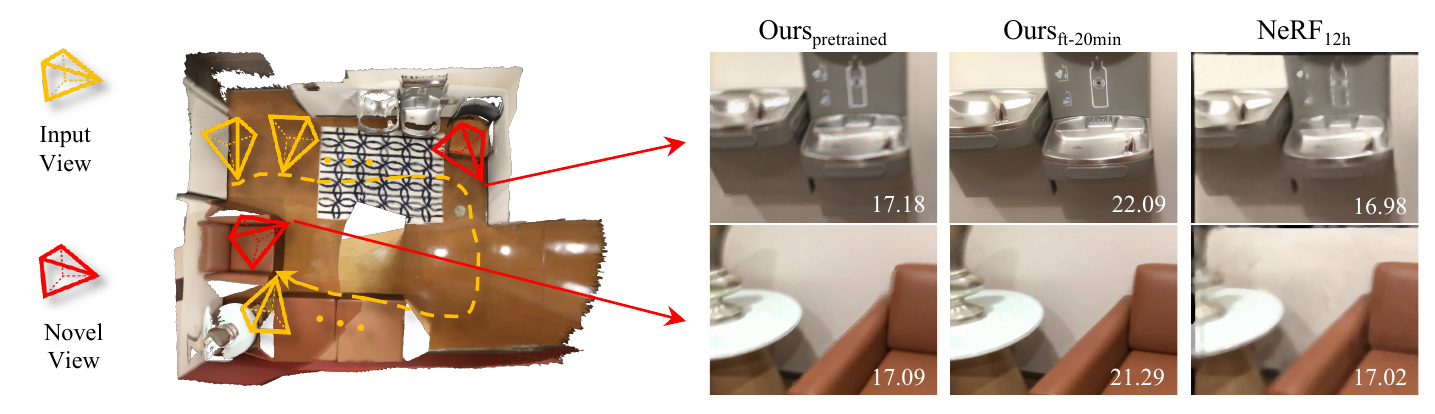}
\captionof{figure}{
 We propose a novel method to enable \emph{fast} reconstruction of volumetric radiance fields of 
\emph{large-scale} scenes. 
Our method uses a novel recurrent network -- that is \emph{generalizable} and trained across scenes -- to sequentially reconstruct a radiance field of a large indoor scene in ScanNet \cite{dai2017scannet} from an input image sequence (marked in yellow) via direct inference.
The predicted field can be directly used to render realistic images at novel viewpoints (marked in red), achieving comparable quality to NeRF~\cite{mildenhall2020nerf} that takes 12 hours per-scene optimization. 
This radiance field can be further fine-tuned for a short period of 20 min, leading to boosted quality that significantly outperforms NeRF.
}
\label{fig:teaser}
\end{strip}

\begin{abstract}
While NeRF \cite{mildenhall2020nerf} has shown great success for neural reconstruction and rendering, its limited MLP capacity and long per-scene optimization times make it challenging to model large-scale indoor scenes.
In contrast, classical 3D reconstruction methods can handle large-scale scenes but do not produce realistic renderings.
We propose \modelname, a method that combines the advantages of NeRF and TSDF-based fusion techniques to achieve efficient large-scale reconstruction and photo-realistic rendering.
We process the input image sequence to predict per-frame local radiance fields via direct network inference. 
These are then fused using a novel recurrent neural network that incrementally reconstructs a global, sparse scene representation in real-time at 22 fps.
This global volume can be further fine-tuned to boost rendering quality. 
We demonstrate that \modelname achieves state-of-the-art quality on both large-scale indoor and small-scale object scenes, with substantially faster reconstruction than NeRF and other recent methods.\footnote{\href{https://jetd1.github.io/NeRFusion-Web/}{https://jetd1.github.io/NeRFusion-Web/}}
\end{abstract}

\section{Introduction}
\label{sec:intro}

Reconstructing and rendering large-scale indoor scenes from RGB images is challenging but crucial for various applications in computer vision and graphics, including AR/VR, e-commerce, and robotics.
While truncated signed distance function (TSDF) fusion techniques \cite{newcombe2011kinectfusion,weder2020routedfusion} can achieve efficient reconstruction, these methods often use depth sensors and focus on geometric reconstruction only, and cannot synthesize realistic images.
Recently, NeRF \cite{mildenhall2020nerf} proposed optimizing scene radiance fields, represented using global MLPs, from RGB images to achieve photo-realistic novel view synthesis. However, NeRF cannot handle large-scale scenes well due to its limited MLP network capacity and impractical slow per-scene optimization.  


In this work, we aim to achieve \textit{fast, large-scale} scene-level radiance field reconstruction to make neural scene reconstruction and rendering more practical. As opposed to small-scale object-centric scenes, we use ``large-scale'' to refer to full-size indoor scenes, like ScanNet scenes~\cite{dai2017scannet}), with multiple rooms and objects with complex scene geometry and appearance.
To achieve fast radiance field reconstruction on such challenging scenes, we propose a novel neural framework that uses recurrent neural modules to incrementally reconstruct a large sparse radiance field from a long RGB image sequence.
Unlike NeRF \cite{mildenhall2020nerf}, that requires per-scene optimization, our network is generalizable, pre-trained across scenes, and able to efficiently reconstruct large-scale radiance fields via direct network inference.
As shown in Fig.~\ref{fig:teaser}, our framework can successfully reconstruct a large indoor scene from from an input monocular RGB video from ScanNet \cite{dai2017scannet}, to create a high-quality radiance field with photo-realistic novel view synthesis results. 

Our reconstructed radiance field is represented by a sparse volume grid with per-voxel neural features; these voxel features are tri-linearly interpolated at any scene location, and used to regress volume density and view-dependent radiance through an MLP decoder for differentiable volume rendering.
In contrast to previous methods \cite{liu2020neural,hedman2021baking} that reconstruct similar representations using slow per-scene optimization, we present a novel deep neural network that can be trained across scenes and generalize on unseen novel scenes to achieve fast radiance field reconstruction, bypassing per-scene fitting.  

Given an input sequence of RGB images with known camera poses (that can be registered by SLAM or SfM techniques),
our framework reconstructs a radiance field as a sparse neural volume.
Our pipeline is inspired by the classical TSDF fusion workflow \cite{newcombe2011kinectfusion,niessner2013real,weder2020routedfusion} that starts from per-view geometry (depth) and fuses the per-view reconstruction across key frames to obtain a global sparse TSDF volume. 
This workflow is widely used to reconstruct large-scale scenes, but only focuses on geometric reconstruction.
Instead, we propose novel neural modules to reconstruct radiance fields as sparse voxels for photo-realistic rendering. 

We first reconstruct local radiance fields for each input key frame. 
We leverage deep MVS techniques and apply sparse 3D convolutions on a world-space cost-volume built from unprojected 2D images features (regressed from a deep 2D CNN) of neighboring key frames.
This reconstructs sparse neural voxels that represent a local radiance field.
Once estimated, this field can already be used to render realistic images locally, though only for partial scene content seen by the local frames.
We propose a recurrent neural fusion module to sequentially fuse multiple local fields across frames.
Our fusion module recurrently takes a newly estimated local field as input and learns to incorporate the local voxels to progressively reconstruct a global radiance field modeling the entire scene, by adding new voxels and updating existing voxels.
Our full model is trained from end to end, learning to reconstruct radiance fields with arbitrary scene scales from an arbitrary number of input images. 
We show that our direct network output can already render high-quality images; moreover, our neural field can be effectively fine-tuned by optimizing the predicted voxel features per scene in a short period to achieve better rendering quality (see Fig.~\ref{fig:teaser} and \ref{fig:results}).

We train our full framework from end to end with only rendering losses on a combination of scenes from the ScanNet \cite{dai2017scannet}, DTU \cite{dtu}, and Google Scanned Object \cite{scannedobjects} datasets. These datasets contain a large variety of different objects and scenes, allowing for our method to work properly with any scene scale.
We demonstrate, on various datasets, that our approach performs better than prior arts, including IBRNet \cite{ibrnet} that also designs networks that generalize across scenes. Especially on large-scale indoor scenes, our results from real-time direct network inference can even be on par with NeRF's results from long per-scene optimization. 
Moreover, after only one hour of per-scene fine-tuning, our quality can be further boosted to the state-of-the-art, outperforming NeRF \cite{martin2021nerf} and NVSF \cite{liu2020neural} that require much longer per-scene optimization times.
Our approach significantly improves the efficiency and scalability of radiance field reconstruction. We believe this is an important step towards making neural scene reconstruction and rendering practical.

\section{Related Work}
\label{sec:rel}
\boldstart{Multi-view scene reconstruction.}
Abundant research has been conducted on reproducing the appearance of of 3D scens 
from multi-view data. To reconstruct the geometry, previous methods 
apply multi-view stereo~\cite{schoenberger2016mvs,seitz2006comparison} or 
depth sensors~\cite{newcombe2011kinectfusion} to acquire the depth information of the 
scene. Recently, learning-based multi-view stereo 
methods~\cite{huang2018deepmvs,yao2018mvsnet,yao2019recurrent,chen2019point,luo2019p,gu2020cascade,cheng2020deep} 
based on plane-swept cost volumes are also introduced for depth estimation.
Given the depth, one category of methods represent scenes with
colored point clouds~\cite{aliev2020neural,kopanas2021point,lassner2021pulsar}, 
and utilize point splatting to render images of the scene. 
Another category of 
methods~\cite{newcombe2011kinectfusion,niessner2013real,weder2020routedfusion} fuse multi-view depth and 
reconstruct surface meshes using techniques such as TSDF fusion or Poisson 
reconstruction, and further generate textures~\cite{zhou2014color,bi2017patch} 
from multi-view images. 
However, both kinds of methods are sensitive to potential
inaccuracies in point clouds and meshes resulting from corrupted depth, especially 
when there are thin structures and textureless regions, thus
suffering from holes and blurry artifacts in the final renderings. 
While some works apply neural networks~\cite{aliev2020neural,ruckert2021adop,hedman2018deep} such as a 2D CNN in screen-space 
to mitigate potential errors in the geometry, their models are per-scene optimized for a specific scene (similar to NeRF), requiring a long optimization time.
Moreover, the screen-space neural networks typically produce temporally unstable 
results with flickering artifacts.
Instead of estimating and fusing per-view depth, previous methods~\cite{lsmKarHM2017,sun2021neucon,bozic2021transformerfusion} 
introduce learning-based methods to aggregate per-view features and 
predict opacity volumes or signed distance volumes. These methods only focus on 
geometry reconstruction and cannot produce realistic renderings. 
In contrast, our approach models scenes as neural volumetric radiance fields and can reproduce the faithful scene appearance, producing photo-realistic novel views.   
Our pipeline is pretrained on multi-view image datasets and can generalize to novel scenes
at arbitrary scales and enable efficient large-scale neural reconstruction. 
\vspace{-0.6em}

\boldstartspace{Neural Radiance Fields.}
Volumetric representations~\cite{lombardi2019neural,mildenhall2020nerf,liu2020neural} 
have been widely adopted to reconstruct the appearance of 
the scene. NeRF~\cite{mildenhall2020nerf} uses a global MLP to regress the 
volume density and view-dependent radiance at any arbitrary point in the space, and applies 
volume rendering to synthesize images at novel viewpoints. Following works extend 
the framework for different tasks such as relighting~\cite{bi2020deep,bi2020neural,boss2021nerd,nerv2021}, 
scene editing~\cite{xiang2021neutex} 
and dynamic scene modeling~\cite{park2021nerfies,park2021hypernerf,li2021neural}. 
Similar to NeRF, most of these works train MLP networks, 
specific for each scene from scratch, which can take hours and even days to optimize, heavily time-consuming. On the other hand, the limited network capacity of MLPs makes these methods hard to scale up to large scene reconstruction.  NVSF~\cite{liu2020neural} improves the scalability by building sparse voxel grids with per-voxel features. However their 
networks and features are still optimized per scene from 
scratch and it can still take days for large-scale scenes.   
In contrast, while we use a similar sparse volume, 
our volume is generated from the direct inference of a pre-trained network, leading to fast large-scale scene reconstruction.
The direct network output can be further fine-tuned in a short period to achieve better rendering quality than NeRF and NSVF.

Some previous papers  
also extend NeRF for generalization.
PixelNeRF \cite{yu2020pixelnerf} uses 2D CNNs to extract image features on each sampled point of each ray used in ray marching for regressing the point's volume properties.
However, their network is designed for object rendering with few images and is trained specifically for each dataset.
IBRNet \cite{ibrnet} uses a similar network but has better designs that enable rendering on any scene scales.
However, it leverages image features from neighboring views as input, varying across novel viewpoints, which often lead to blurry or flickering artifacts from sparse inputs. 
Our network instead reconstructs a neural volume with per-voxel features in 3D, modeling scene geometry and appearance in a more consistent way, leading to much better rendering than IBRNet.
MVSNeRF \cite{chen2021mvsnerf} also reconstructs 3D volumes; however, it focuses on reconstructing a local volume from a fixed number of three nearby views. In contrast, our network can fuse per-view local reconstruction into a global volume from an arbitrary number of images, leading to highly efficient large-scale scene reconstruction and rendering.


\section{Method}
\label{sec:method}

\newcommand{\comment}[1]{}

\newcommand{\V}{\mathcal{V}}
\newcommand{\U}{\mathcal{U}}
\newcommand{\VoxC}{v}
\newcommand{\VoxP}{u}

\newcommand{\Img}{I}
\newcommand{\Fea}{F}
\newcommand{\Cam}{\Phi}
\newcommand{\iI}{t}
\newcommand{\iIN}{i}

\newcommand{\ImgNum}{N}
\newcommand{\LocalNum}{K}
\newcommand{\LocalHood}{\mathcal{N}}

\newcommand{\Color}{c}
\newcommand{\ShadX}{x}
\newcommand{\iS}{j}
\newcommand{\iSS}{t}
\newcommand{\ShadNum}{M}

\newcommand{\Dir}{d}
\newcommand{\Rad}{c}
\newcommand{\Trans}{\tau}
\newcommand{\Dens}{\sigma}
\newcommand{\Step}{\Delta}

\newcommand{\NetVol}{R}

\newcommand{\NetDir}{G}
\newcommand{\NetLocal}{J}
\newcommand{\NetZ}{M_z}
\newcommand{\NetR}{M_r}
\newcommand{\NetT}{M_t}

\begin{figure*}[t]\centering
    \includegraphics[width=\textwidth]{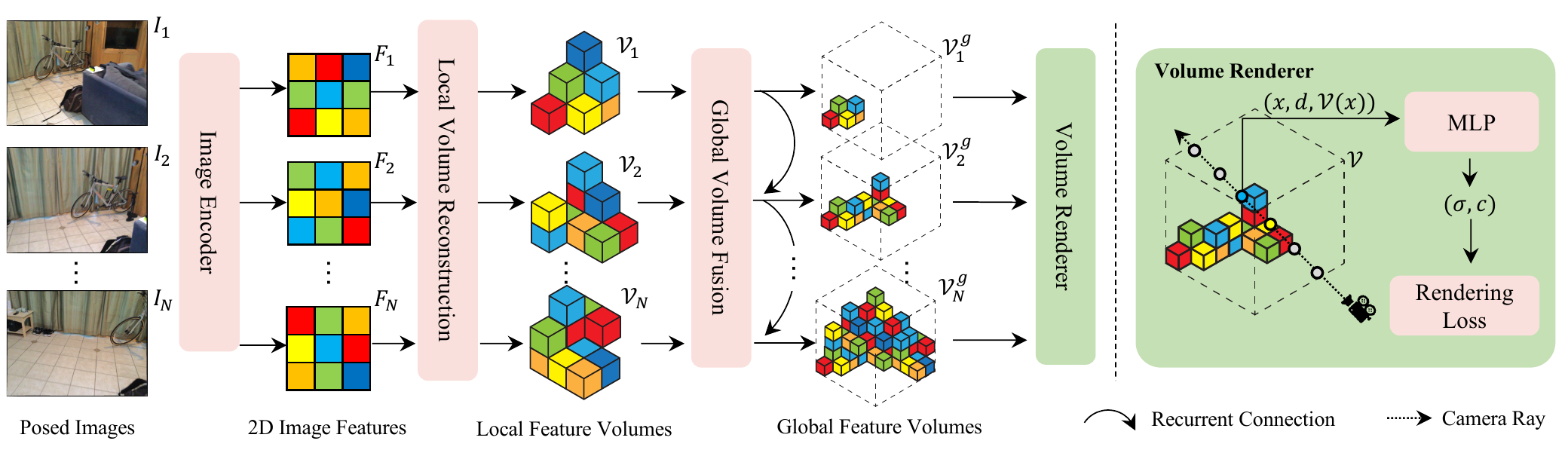}
    \caption{
        Overview of our framework. Given a sequence of images, 1) we first extract their image features ($F_1 \dots F_\ImgNum$) using a 2D CNN. 2) Then,
        at each frame, we reconstruct a local sparse neural volume $\V_1 \dots \V_\ImgNum$ in the canonical world space by fetching and aggregating 2D features across its neighboring views at visible voxels using a sparse 3D CNN.   
        3) We further 
        fuse the local sparse volumes across frames using a recurrent neural network and sequentially build global feature volumes $\V_1^g \dots \V_\ImgNum^g$ to model a radiance field of the entire scene. 
        We regress volume density and view-dependent radiance from the sparse neural volumes to render images with differentiable ray marching.
    }
    \vspace{-1.5em}
    \label{fig:pipeline}
\end{figure*}

We now present our approach for neural scene reconstruction and rendering.
Given an input sequence of images $\Img_1$,...,$\Img_\ImgNum$ of a large-scale scene with their known camera parameters $\Cam_1$,...,$\Cam_\ImgNum$, our approach reconstructs a radiance field
modeling the entire scene for realistic rendering.


Our final output radiance field is represented by a sparse neural volume $\V^g$ with per-voxel neural features
(Sec.~\ref{sec:representation}).
Unlike per-scene optimization methods \cite{mildenhall2020nerf,liu2020neural}, we propose a deep neural network that sequentially takes the images $\Img_\iI$ frame by frame as input and convert the sequence to the final sparse reconstruction via direct network inference.
Our pipeline first learns to reconstruct a sparse volume $\V_\iI$ per input image frame, expressing a local radiance field covered by local frames (Sec.~\ref{sec:localrecon}).
We then leverage a recurrent fusion module that learns to fuse the per-frame volumes $\V_\iI$ online, incrementally reconstructing the global large-scale field $\V^g$ (Sec.~\ref{sec:fusion}).
We train our full pipeline from end to end with pure rendering losses. Our model can reconstruct high-quality radiance fields from direct network inference; the estimated field can also be further fine-tuned to boost its quality (Sec.~\ref{sec:training}).

\subsection{Sparse Volumes for Radiance Fields}
\label{sec:representation}



Our output radiance field is modeled by a sparse neural volume $\V$ that has per-voxel neural features in voxels that approximately cover the actual scene surface. 
We regress volume density $\Dens$ and view-dependent radiance $\Rad$ at any given 3D location $\ShadX$ from this volume using an MLP network, in which we first tri-linearly sample a feature vector and then use the MLP to convert the feature to volume properties, expressed by 
\begin{align}
\label{eq:radiancefield}
\Dens, \Rad &= \NetVol(\ShadX, \Dir, \V(\ShadX)).
\end{align}
Here $\V(\ShadX)$ represents the trilinearly interpolated feature at $\ShadX$, $\NetVol$ is the MLP, and $d$ is the viewing direction in rendering. 
The output volume properties regressed from the volume can be directly used to synthesize images at novel target viewpoints via differentiable ray marching as is done in NeRF \cite{mildenhall2020nerf}. 
This radiance field representation is similar to the one in Neural Sparse Voxel Fields \cite{liu2020neural} that purely relies on per-scene optimization for the reconstruction. 
We instead propose to leverage neural networks trained across scenes to predict the neural volumes from image sequences.

In our pipeline, such sparse volumes are reconstructed locally as $\V_\iI$ per frame $\iI$ and also globally as $\V_g$ for the entire sequence. The MLP network $\NetVol$ is shared across all volumes in the training process. We model the local volumes and the global volume both in the canonical world space. 
 

\subsection{Reconstructing Local Volumes}
\label{sec:localrecon}
We propose a deep neural network to regress a local neural volume for each input frame $\iI$, using its image $\Img_\iI$ and $\LocalNum-1$ images from neighboring views. Usually, given a monocular video, these neighboring views correspond to temporal neighboring frames. 
Using multiple nearby images for per-frame reconstruction allows the network to leverage multi-view correspondence to recover better scene geometry, which a single image cannot provide.

To make the local reconstruction per frame well generalized across scenes, we leverage deep MVS techniques \cite{yao2018mvsnet,ji2017surfacenet}, which are known to be generalizable. 
We extract 2D image features, build a cost volume from the features, and regress a neural feature volume from the cost volume.
However, unlike MVSNeRF \cite{chen2021mvsnerf} and other MVS techniques \cite{yao2018mvsnet,cheng2020deep} that built frustum volumes in view's perspective coordinate, we construct volumes in the canonical world coordinate frame to align it with the final global volume output $\V_g$, facilitating the following fusion process.
\vspace{-0.5em}



\boldstartspace{Image feature extraction.} 
We use a deep 2D convolutional neural network to extract 2D image features for each input image. This network maps the input image $\Img_\iI$ into a 2D feature map $\Fea_\iI$, encoding the scene content from each view.
\vspace{-1.5em}

\boldstartspace{Local sparse volume.} 
We consider the bounding box that covers the frustums of all $\LocalNum$ neighboring viewpoints in the world coordinate frame, containing of a set of voxels in the canonical space.
The bounding volume is axis-aligned with the world frame; each voxel inside it can be visible to a different number of neighboring views. We mask out all the voxels invisible to all view, leading to a sparse set of voxels in the bounding box.
We then unproject the image features into this volume for our local reconstruction.
\vspace{-0.3em}

\boldstartspace{3D feature volume.} 
For each neighboring viewpoint $\iIN$ and its feature map $\Fea_\iIN$, we build a 3D feature volume $\U_\iIN$.
In particular, for each visible voxel centered at $\VoxC$, we fetch the 2D image feature at its 2D projection from each neighboring view at frame $\iI$. In addition to pure image features, we leverage the corresponding viewing direction $\Dir_\iIN$ at $\VoxC$ from each viewpoint and compute additional features using an MLP $\NetDir$.
The per-view 3D volume $\U_\iIN$ is expressed by
\begin{align}
\U_\iIN(\VoxC) = [\Fea_\iIN(\VoxP_\iIN), \NetDir(\Dir_\iIN)],
\end{align}
where $\U_i(\VoxC)$ is the feature at a voxel centered at $\VoxC$, $\VoxP_\iIN$ is the center's 2D projection in view $\iIN$, $[\cdot, \cdot]$ represents feature concatenation.
Note that, we encode the additional information of input viewing directions in the reconstruction process; this crucial information makes our following fusion module effectively account for the view-dependent effects captured across frames.
\vspace{-0.6em}

\boldstartspace{Neural reconstruction.} 
We then aggregate the features across multiple neighboring viewpoints to regress a local volume $\V_\iI$ at frame $\iI$, expressing a local radiance field.
We propose to leverage the mean and variance of the per-voxel features in $\U_i$ computed across neighboring viewpoints; such operations have been widely used in building cost volumes in MVS-based techniques \cite{yao2018mvsnet,chen2021mvsnerf}, where the mean can fuse per-view appearance information and the variance provides rich correspondence cues for geometry reasoning. 
These two operation are also invariant to the number/order of input; in our case, this naturally handles voxels that have different numbers of visible viewpoints.
We use a deep neural network $J$ to process the mean and variance features per voxel to regress the per-view reconstruction by
\begin{align}
\V_t &= J([\text{Mean}_{\iIN \in \LocalHood_\iI}\U_\iIN,\ \text{Var}_{i \in \LocalHood_\iI}\U_\iIN]),
\end{align}
Here $\LocalHood_\iI$ represents all $\LocalNum$ neighboring viewpoints used at frame $\iI$; $\text{Mean}$ and $\text{Var}$ represent element-wise average and variance operation, respectively.

Essentially, we regress the local radiance field from the features across neighboring views. This is similar to MVSNeRF \cite{chen2021mvsnerf}. However, unlike MVSNeRF that considers only local reconstruction and builds perspective frustrum volumes for small-baseline rendering, we leverage these local volumes for global large-scale reconstruction and rendering. We build volumes directly in canonical space, naturally providing per-frame voxel inputs for our fusion module.




\begin{figure}\centering
    \includegraphics[width=0.42\textwidth]{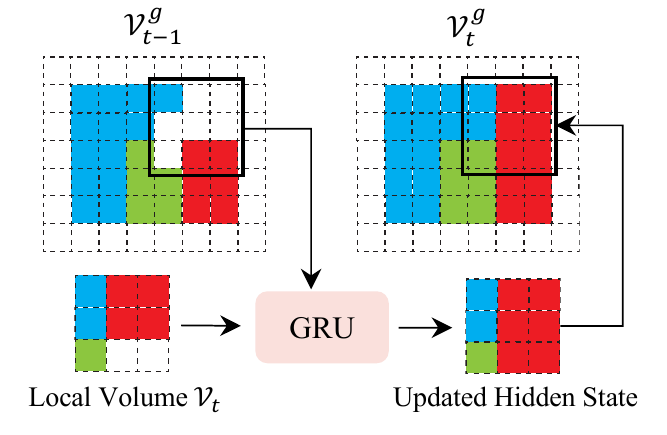}
    \captionof{figure}{A 2D example illustrating the GRU fusion step. The hidden state, which is also the global feature volume $\V^g_{\iI-1}$, is adaptively updated by aggregating new information in the incoming local feature volume $\V_\iI$.}
    \vspace{-1.4em}
    \label{fig:ops}
\end{figure}

\subsection{Fusing Volumes for Global Reconstruction}
\label{sec:fusion}

In order to create a consistent, efficient, and extensible scene reconstruction, we propose to use a global neural volume fusion network to incrementally fuse local feature volumes $\{\V_\iI\}$ per frame into a global volume $\V^g$. 

\boldstartspace{Fusion.}
At each frame $\iI$, we consider its local sparse volume reconstruction $\V_\iI$ and the global reconstruction $\V^g_{\iI-1}$ from the previous frame as recurrent input. 
We leverage GRUs (Gated Recurrent Unit)~\cite{cho2014learning} with sparse 3D CNNs in our fusion module, allowing our network to learn to recurrently fuse the per-frame local reconstruction and output high-quality global radiance fields.
This is expressed by
\begin{align}
    z_\iI &= \NetZ([\V^g_{\iI-1}, \V_\iI]), \\
    r_\iI &= \NetR([\V^g_{\iI-1}, \V_\iI]), \\
    \tilde{\V}^g_{\iI} &= \NetT([r_\iI * \V^g_{\iI-1}, \V_\iI]), \\
    \V^g_{\iI} &= (1-z_\iI) * \V^g_{\iI-1} + z_\iI * \tilde{\V}^g_{\iI},
\end{align}
where $*$ is the element-wise multiplication, $z_\iI$ and $r_\iI$ are the update gate and the reset gate, $\NetZ$, $\NetR$ and $\NetT$ all deep neural networks with sparse 3D convolution layers. As in standard GRU, $\NetZ$ and $\NetR$ are designed with sigmoid activation in the end, while $\NetT$ uses tanh, allowing for the entire model sequentially updating the global reconstruction $\V^g_{\iI}$ (seen as the hidden state in a GRU) for every input frame.
In this process, we only apply the networks on the voxels covered by the local volume $\V_\iI$; all other voxels in the global volume are kept unchanged. A 2D illustration of this GRU fusion process is shown in Fig.~\ref{fig:ops}.

Intuitively, the update gate $z_\iI$ and reset gate $r_\iI$ in the GRU determine how much information from the previous global volume $\V^g_{\iI-1}$ as well as how much information from the current local volume $\V_\iI$ should be incorporated into the new global features. In this way, our module can adaptively improve the global scene reconstruction by filling up holes and refining features while keeping the representation consistent. This fusion process is similar to previous 3D reconstruction pipelines~\cite{pizzoli2014remode,kar2017learning,sun2021neucon} that focus on geometry reconstruction; in contrast, we instead reconstruct neural feature volumes to represent neural radiance fields for volume rendering, leading to photo-realistic novel view synthesis.
\vspace{-0.5em}

\boldstartspace{Voxel pruning.}  
To maximize the memory and rendering efficiency, we adaptively prune the global volume reconstruction $\V^g_\iI$ for every frame by removing the non-essential voxels that do not have any scene content inside. We naturally leverage the volume density in each voxel regressed by our radiance field (Eqn.~\ref{eq:radiancefield}), which models the scene geometry. In particular, we prune voxels $V$ if:

\begin{equation}
    \min_{i=1...k} \exp(-\sigma(\VoxC_i)) > \gamma, \VoxC_i \in V, \label{eqn:prune}
\end{equation}

where $\{\VoxC_i\}_{i=1}^k$ are $k$ uniformly sampled points inside the voxel $V$, $\Dens(\VoxC_i)$ is the predicted density at location $\VoxC_i$, and $\gamma$ is a pruning threshold. This pruning step is performed in both later training phase and inference phase once we get a global feature volume $\V^g_\iI$. 
By doing so, we make our global volume sparser, leading to more efficient reconstruction and rendering.

\subsection{Training and optimization}
\label{sec:training}




Once a radiance field (that is represented by a sparse neural volume as described in Sec.~\ref{sec:representation}) is reconstructed, our final rendering is achieved via differentiable ray marching using the regressed volume density and view-dependent radiance at any sampled ray points, as is done in NeRF and any other radiance field methods \cite{mildenhall2020nerf,liu2020neural}. 
In this work, our full pipeline is trained, completely depending on the rendering supervision with the ground truth images, without any extra geometry supervision.

In particular, we first train our local reconstruction network and the radiance field decoder (\NetVol) with a loss 
\begin{equation}
    \mathcal{L}_\text{local} = \|C_\iI -\hat{C} \|^2_2, \label{eq:losslocal}
\end{equation}
where $\hat{C}$ is the ground truth pixel color and $C_\iI$ represents the rendered pixel color using the local volume $\V_\iI$ reconstructed at frame $\iI$.
This makes the network learn to predict reasonable local neural volumes, which are already renderable and able to produce realistic images locally; it also initializes the radiance field decoder MLP to a reasonable state, which is later shared across local and global volumes.
This pre-training allows the local reconstruction module to provide meaningful volume features for the fusion module to utilize in the end-to-end training, effectively facilitating the fusion task.
We then train our full pipeline with the local reconstruction network, fusion network, and the radiance field decoder network all together from end to end, using a rendering loss:
\begin{equation}
    \mathcal{L}_\text{fuse} = \sum_\iI \|C_\iI - \hat{C} \|^2_2 + \|C^g_\iI - \hat{C} \|^2_2 ,
\end{equation}
where $C_\iI$ is the pixel color rendered from the local reconstruction $\V_\iI$ (as is in Eqn.~\ref{eq:losslocal}) and $C^g_\iI$ is the color rendered from the global volume $\V^g_\iI$ after fusing frame $\iI$.  
Basically, we take every intermediate global and local volume ($\V_\iI$ and $\V^g_\iI$) at every frame to render novel view images and supervise them with the ground truth.
The fusion module thus reasonably learns to fuse local volumes from an arbitrary number of input frames.

After trained, our full network is able to output a high-quality radiance field from direct network inference and produce realistic rendering results (as shown in Fig.~\ref{fig:teaser}).
In addition, the reconstructed radiance field as a sparse neural volume can also be easily optimized (fine-tuned) per scene further to boost the rendering quality. 
\vspace{-0.5em}


\boldstartspace{Fine-tuning.}
To fine-tune the estimated radiance field, we optimize the per-voxel neural features in the sparse volume reconstruction $\V^g$ and the MLP decoder per scene with the captured images, leading to better rendering results.
Since our initial reconstruction is already very good, a short period of optimization with less than 25k iterations can usually lead to very high quality, which takes less than 1 hour.
This is substantially less optimization time than NeRF and other pure per-scene optimization methods.

In this per-scene optimization stage, we also do a coarse-to-fine reconstruction, similar to NSVF \cite{liu2020neural}.
Basically, after every 10k optimization iterations, we further prune unnecessary voxels (using Eqn.~\ref{eqn:prune}) and also subdivide each voxel into 8 sub voxels. This prune and subdivision step progressively increases the spatial resolution of the neural volume, further improving our final rendering quality.

\section{Implementation Details.}
\label{sec:impl}

\boldstart{Training datasets.} 
Our training data consists of both large-scale indoor scenes from ScanNet~\cite{dai2017scannet} and small objects from DTU~\cite{jensen2014large} and Google Scanned Objects~\cite{scannedobjects}. 
We randomly sample 100 scenes from ScanNet for training. 
For DTU, We adopt the training split from PixelNeRF~\cite{yu2020pixelnerf}, which includes 88 training scenes. 
In addition, we use the object-centric synthetic renderings of 1,023 models from the Google Scanned Objects~\cite{scannedobjects} generated by~\cite{ibrnet}. 
Our training data includes various camera setups and scene types, enabling our model to generalize to all kinds of scenarios.
We demonstrate that our model can effectively work on large-scale indoor scenes, as well as scenes of objects.

\boldstartspace{Training details.} For each input image sequence, we uniformly sample key frames from the full sequences for the input frames in network training. For object-centric datasets, we sample 16 views for each scene, and for ScanNet, we sample $2\% - 5\%$ of the full sequence as key frames. All other frames are used for supervision.
We use $\LocalNum=3$ neighboring views for each input frame, for local volume reconstruction. 
For video sequence captured by a monocular camera, such as the scenes in ScanNet, we directly take the 3 neighboring key frames temporally. For other datasets, we select the 3 spatially closest viewpoints, in terms of both viewing location and direction.

The sparse volumes and networks are implemented with torchsparse~\cite{tang2020searching}. We train our model using Adam optimizer with an initial learning rate of $0.003$. We train our network with 2 NVIDIA 2080Ti GPUs for 3 days.
During inference,
our network processes frames from ScanNet sequences in real-time at 22 FPS. The final model takes 38 seconds on average to render a $640 \times 480$ image on ScanNet .

\section{Results}
\label{sec:exp}

In this section, we evaluate the our model on various datasets.
For all results, we denote our results from direct network inference as Ours and our results after per-scene fine-tuning as Ours\textsubscript{ft} in all figure and tables. Similar labels are applied to IBRNet and other generalizing methods.
\vspace{-0.5em}

\boldstartspace{Baselines.}
We compare our method against the state-of-the-art NeRF methods on novel view synthesis including per-scene optimization methods, such as NeRF~\cite{mildenhall2020nerf}, NVSF~\cite{liu2020neural}, and NerfingMVS~\cite{wei2021nerfingmvs}, and methods that can generalize to new scenes, such as PixelNeRF~\cite{yu2020pixelnerf}, IBRNet~\cite{ibrnet}, and MVSNeRF~\cite{chen2021mvsnerf}.

\begin{figure*}\centering
    \includegraphics[width=\textwidth]{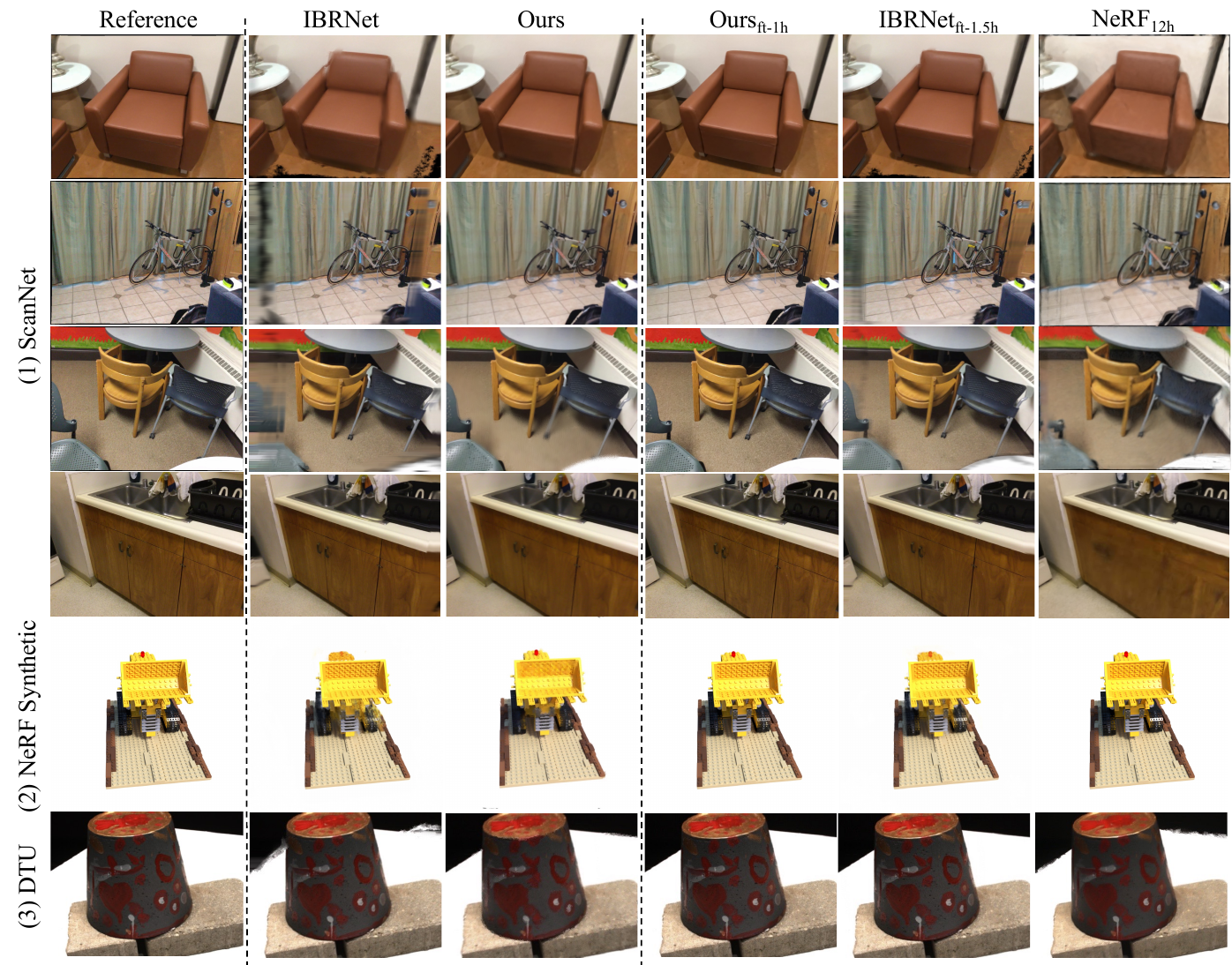}
    \captionof{figure}{
        Qualitative comparisons of rendering quality on diverse scenes between our method and state-of-the-art 
        method. Our method achieves better performance than state-of-the-art 
        generalizable method IBRNet~\cite{ibrnet} in both the direct-inference 
        and the fine-tuned settings, where 
        IBRNet generates results with obvious blurry and tearing artifacts. Our model fine-tuned 
        for 1 hour can generate even better results than NeRF~\cite{mildenhall2020nerf} that requires 12 hours of training, especially 
        on large-scale scenes from ScanNet~\cite{dai2017scannet}.
    }
    \vspace{-1.5em}
    \label{fig:results}
\end{figure*}

To achieve fair and accurate comparisons, we run our method on the same experiment settings in previous papers, and we try our best to directly use the reported official quantitative results in previous papers or use the official code to run the experiments.
We find that the official NeRF and IBRNet code can easily run and work on different datasets, producing corresponding images.
We demonstrate visual comparison with these two methods across the three testing sets in Fig.~\ref{fig:results}.
On the other hand, NSVF is very hard to run without enough GPU memory; their official models are optimized on a V100 GPU that has 32G memory; we found it impractical to generate their corresponding results with our resources. We therefore only include NSVF's quantitative results whenever they are reported previously.
Besides, the recent MVSNeRF~\cite{chen2021mvsnerf} is a very relevant technique, but it is designed to take a fixed number of three nearby views as its network input; as a result, it cannot support large-baseline rendering or arbitrary number of input images for inference. We therefore only compare with MVSNeRF on the DTU dataset in the same experiment setting used in their paper.
\vspace{-0.8em}

\begin{table}
    \centering
    \setlength{\tabcolsep}{4.0pt}
    \begin{tabular}{@{}lcccc@{}}
        \toprule
        Method      & Settings & PSNR$\uparrow$ & SSIM$\uparrow$ & LPIPS$\downarrow$  \\
        \midrule
        IBRNet                          & \multirow{2}{*}{\shortstack{No per-scene \\ optimization}}                  &{21.19}      &{0.786}       &{0.358}          \\
        Ours                            &              &\B{{22.99}}  &\B{{0.838}}   &\B{{0.335}}     \\
        \midrule
        NeRF                            &  \multirow{5}{*}{\shortstack{Per-scene \\ optimization}}              &24.04        &0.860         &0.334    \\
        NSVF                            &            &26.01        &0.881         &-         \\
        NeRFingMVS                      &            &26.37        &0.903         &0.245     \\
        IBRNet\textsubscript{ft-1.5h}   &            &{25.14}      &{0.871}       &{0.266}    \\
        Ours\textsubscript{ft-1h}       &            &\B{{26.49}}  &\B{{0.915}}   &\B{{0.209}}   \\
        \bottomrule
    \end{tabular}\vspace{-0.5em}
    \caption{
        Quantitative comparisons on the ScanNet dataset~\cite{dai2017scannet}. We follow the same experiment settings 
        as in NeRFingMVS~\cite{wei2021nerfingmvs} and report the error metrics including PSNR (higher is better), SSIM (higher is better) and 
        LPIPS (lower is better). 
        Note that, our diret inference results are better than IBRNet~\cite{ibrnet}.
        Our fine-tuning results achieve the best numbers in all three metrics.
    }
    \label{tab:scannet8}
\end{table}



\begin{table}
    \centering
    \setlength{\tabcolsep}{4.0pt}
    \begin{tabular}{@{}lcccc@{}}
        \toprule
        Method      & Settings & PSNR$\uparrow$ & SSIM$\uparrow$ & LPIPS$\downarrow$      \\
        \midrule
        IBRNet          & \multirow{2}{*}{\shortstack{No per-scene \\ optimization}}                                 &\B{25.51}  &0.916        &0.100         \\
        Ours       &                                    &\B{{25.47}}    &\B{{0.922}}  &\B{{0.093}}   \\
        \midrule
        NeRF         & \multirow{4}{*}{\shortstack{Per-scene \\ optimization}}                                    &31.01      &0.947        &0.081       \\
        NSVF         &                                 &\B{31.75}  &\B{0.954}    &\B{0.048}   \\
        IBRNet\textsubscript{ft-1.5h}  &                &28.19      &0.943        &0.072       \\
        Ours\textsubscript{ft-1h}  &               &{31.25}    &\B{{0.953}}  &{0.069}     \\
        \bottomrule
    \end{tabular}\vspace{-0.5em}
    \caption{
        Quantitative comparison on the NeRF Synthetic dataset~\cite{mildenhall2020nerf}.
        Our model is able to generate better results than IBRNet in both direct inference 
        and fine-tuning settings. Our model after 1 hour fine-tuning achieves comparable performance
        to the state-of-the-art per-scene overfitting methods such as NeRF~\cite{mildenhall2020nerf}
        and NVSF~\cite{liu2020neural}. 
    }\vspace{-1.2em}
    \label{tab:nerfsynth}
\end{table}

\boldstartspace{Large-scale scenes in ScanNet.} 
We follow the same training and evaluation scheme as described in
NerfingMVS~\cite{wei2021nerfingmvs} for the comparison on ScanNet. 
We tested our model on the $8$ testing scenes used in their paper\comment{, where each scene contains around 32 training and 8 testing views}.  
From Table~\ref{tab:scannet8}, we can see that our recurrent neural reconstruction network generates significantly
better results than IBRNet via direct network inference.
After fine-tuning for only a short period of 1 hour, the quality of our results is further boosted significantly, leading to the best PSNR, SSIM and LIPIPS in all compared methods. 
Note that, the per-scene optimization methods like NeRF, NeRFing MVS and NSVF require substantially longer per-scene optimization time but are still outperformed by our method.
Our approach is impressively better than NSVF in this case though both methods have similar final radiance field representation; this indicates that the data priors learned by our recurrent neural network can effectively help the reconstruction and lead to reasonable initial radiance fields, even benefiting the per-scene fine-tuning process.

As shown in Fig.~\ref{fig:results}, our results on these large-scale scenes are of very high visual quality.  Our results are visually much better than the IBRNet's results from both direct inference and per-scene fine-tuning. 
IBRNet generates tearing artifacts since it performs image-based rendering and
can only aggregate a small set of local neighboring views due to limited GPU
memory. In contrast, our model learns a unified 3D representation in the canonical space with a recurrent
module that is able to efficiently aggregate per-view information across all input views, leading to significantly better rendering quality with better across-view consistency.
Note that, even our direct inference renderings are already very realistic and contain few noticeable artifacts; they are arguably comparable to the rendering results of NeRF which require long per-scene optimization. 
Our approach achieves highly efficient and highly accurate large-scale radiance field reconstruction.
\vspace{-0.5em}

\begin{table}
    \centering
    \setlength{\tabcolsep}{2.5pt}
    \begin{tabular}{@{}lcccc@{}}
        \toprule
        Method      & Settings & PSNR$\uparrow$ & SSIM$\uparrow$ & LPIPS$\downarrow$  \\
        \midrule
        PixelNeRF       & \multirow{4}{*}{\shortstack{No per-scene \\ optimization}}                             &19.31         &0.789      &0.382    \\
        IBRNet &                                      &26.04         &0.917      &0.190   \\
        MVSNeRF &                                    &\B{26.63}     &\B{0.931}  &\B{0.168}   \\
        Ours     &                                   &{26.19}       &{0.922}    &{0.177}   \\
        \midrule
        NeRF            &  \multirow{4}{*}{\shortstack{Per-scene \\ optimization}}                              &27.01         &0.902      &0.263   \\
        IBRNet\textsubscript{ft-1.5h}       &        &31.35         &0.956      &0.131   \\
        MVSNeRF\textsubscript{ft-15min}      &       &28.50         &0.933      &0.179   \\
        Ours\textsubscript{ft-1h}             &      &\B{{31.79}}   &\B{0.962}  &\B{0.119}   \\
        \bottomrule
    \end{tabular}\vspace{-0.5em}
    \caption{
        Quantitative comparisons on the DTU dataset~\cite{dtu}. Our model is able to generate good results under this difficult setting where only 3 input views are given for the direct inference. Our fine-tuning results outperforms other methods in all three metrics.
    }\vspace{-1.2em}
    \label{tab:dtu3}
\end{table}

\boldstartspace{NeRF Synthetic.} 
Our method also works well on small-scale scenes.
We conduct experiments on the NeRF Synthetic $360^{\circ}$ 
dataset, and apply the same evaluation setting as in~\cite{mildenhall2020nerf}. 
As shown in Table~\ref{tab:nerfsynth}, without per-scene fine-tuning, our model generates results that are 
comparable to IBRNet; however, fine-tuning significantly boosts the performance 
of our model, leading to high accuracy that is much superior to the fine-tuned IBRNet. In practice, IBRNet 
suffers from the sparsely distributed input views of the dataset with large baselines, where 
interpolating neighboring views are not effective to synthesize realistic novel view images. 
Our fine-tuned 
model also achieves similar performance when compared to per-scene optimization 
methods \cite{mildenhall2020nerf,liu2020neural}, while ours is optimized for only 1 hour, substantially less than the time other methods require.
\vspace{-0.5em}

\boldstartspace{DTU.} To show that our method works with a small number of input views with small baselines. We also evaluate our model on the DTU dataset, following the experiment settings in MVSNeRF \cite{chen2021mvsnerf}, where only 3 views are provided for the setting without per-scene optimization and 16 more views are provided in the per-scene optimization setting. Here we also compare with PixelNeRF \cite{yu2020pixelnerf}, which is specifically trained for the DTU dataset in their paper.
As demonstrated in Tab.~\ref{tab:dtu3} and Fig.~\ref{fig:results}, similar to previous results, 
our model generalizes well to the testing scenes and can be efficiently fine-tuned to 
outperform NeRF and other generalizable methods including IBRNet~\cite{ibrnet} and MVSNeRF~\cite{chen2021mvsnerf}.

\section{Limitations.}
\vspace{-0.5em}
\label{sec:slimit}
Our approach currently focuses on handling large-scale indoor scenes, but might not be efficient on handling scenes that have foreground objects with distant background, which might appear in unbounded outdoor scenes. This is because we consider a uniform grid for the entire scene, similar to~\cite{liu2020neural}. This can be potentially addressed in the future by doing per-view reconstruction in disparity space or applying spherical coordinates for regions at long distances (similar to ~\cite{zhang2020nerf++}).
Our method relies on multi-view correspondence; hence, extreme camera poses without enough parallax could lead to problems, which cannot be addressed by any MVS-based techniques. For our current pipeline, we simply sample input frames uniformly, because the camera motion in ScanNet has enough translation. However, a more careful input view selection technique that accounts for relative camera poses may be necessary in practice to address various types of camera motions.

\vspace{-0.3em}
\section{Conclusion}
\vspace{-0.5em}
\label{sec:conclu}
In this work, we present a novel neural approach that can achieve fast, large-scale, and high-quality scene reconstruction for photo-realistic rendering.
In contrast to traditional TSDF-based reconstruction, we reconstruct scenes as volumetric radiance fields, leading to photo-realistic view synthesis results.
Our approach leverages a novel recurrent neural network to process the input image sequence and incrementally reconstruct a global large-scale radiance field by reconstructing and fusing per-frame local radiance fields.
We demonstrate that our approach can achieve the state-of-the-art rendering quality for large-scale indoor scenes from ScanNet while taking substantially less reconstruction time.


{\small
\bibliographystyle{ieee_fullname}
\bibliography{egbib}

\begin{thebibliography}{10}\itemsep=-1pt

\bibitem{aliev2020neural}
Kara-Ali Aliev, Artem Sevastopolsky, Maria Kolos, Dmitry Ulyanov, and Victor
  Lempitsky.
\newblock Neural point-based graphics.
\newblock In {\em Computer Vision--ECCV 2020: 16th European Conference,
  Glasgow, UK, August 23--28, 2020, Proceedings, Part XXII 16}, pages 696--712.
  Springer, 2020.

\bibitem{bi2017patch}
Sai Bi, Nima~Khademi Kalantari, and Ravi Ramamoorthi.
\newblock Patch-based optimization for image-based texture mapping.
\newblock {\em ACM Transaction on Graphics}, 36(4):106--1, 2017.

\bibitem{bi2020neural}
Sai Bi, Zexiang Xu, Pratul Srinivasan, Ben Mildenhall, Kalyan Sunkavalli,
  Milo{\v{s}} Ha{\v{s}}an, Yannick Hold-Geoffroy, David Kriegman, and Ravi
  Ramamoorthi.
\newblock Neural reflectance fields for appearance acquisition.
\newblock {\em arXiv preprint arXiv:2008.03824}, 2020.

\bibitem{bi2020deep}
Sai Bi, Zexiang Xu, Kalyan Sunkavalli, Milo{\v{s}} Ha{\v{s}}an, Yannick
  Hold-Geoffroy, David Kriegman, and Ravi Ramamoorthi.
\newblock Deep reflectance volumes: Relightable reconstructions from multi-view
  photometric images.
\newblock In {\em Proc.~ECCV}, 2020.

\bibitem{boss2021nerd}
Mark Boss, Raphael Braun, Varun Jampani, Jonathan~T Barron, Ce Liu, and Hendrik
  Lensch.
\newblock Nerd: Neural reflectance decomposition from image collections.
\newblock In {\em Proceedings of the IEEE/CVF International Conference on
  Computer Vision}, pages 12684--12694, 2021.

\bibitem{bozic2021transformerfusion}
Alja{\v{z}} Bo{\v{z}}i{\v{c}}, Pablo Palafox, Justus Thies, Angela Dai, and
  Matthias Nie{\ss}ner.
\newblock Transformerfusion: Monocular rgb scene reconstruction using
  transformers.
\newblock {\em Proc. Neural Information Processing Systems (NeurIPS)}, 2021.

\bibitem{chen2021mvsnerf}
Anpei Chen, Zexiang Xu, Fuqiang Zhao, Xiaoshuai Zhang, Fanbo Xiang, Jingyi Yu,
  and Hao Su.
\newblock Mvsnerf: Fast generalizable radiance field reconstruction from
  multi-view stereo.
\newblock {\em arXiv preprint arXiv:2103.15595}, 2021.

\bibitem{chen2019point}
Rui Chen, Songfang Han, Jing Xu, and Hao Su.
\newblock Point-based multi-view stereo network.
\newblock In {\em Proc.~ICCV}, 2019.

\bibitem{cheng2020deep}
Shuo Cheng, Zexiang Xu, Shilin Zhu, Zhuwen Li, Li~Erran Li, Ravi Ramamoorthi,
  and Hao Su.
\newblock Deep stereo using adaptive thin volume representation with
  uncertainty awareness.
\newblock In {\em Proceedings of the CVPR}, pages 2524--2534, 2020.

\bibitem{cho2014learning}
Kyunghyun Cho, Bart Van~Merri{\"e}nboer, Caglar Gulcehre, Dzmitry Bahdanau,
  Fethi Bougares, Holger Schwenk, and Yoshua Bengio.
\newblock Learning phrase representations using rnn encoder-decoder for
  statistical machine translation.
\newblock {\em arXiv preprint arXiv:1406.1078}, 2014.

\bibitem{dai2017scannet}
Angela Dai, Angel~X. Chang, Manolis Savva, Maciej Halber, Thomas Funkhouser,
  and Matthias Nie{\ss}ner.
\newblock Scannet: Richly-annotated 3d reconstructions of indoor scenes.
\newblock In {\em Proc. Computer Vision and Pattern Recognition (CVPR), IEEE},
  2017.

\bibitem{gu2020cascade}
Xiaodong Gu, Zhiwen Fan, Siyu Zhu, Zuozhuo Dai, Feitong Tan, and Ping Tan.
\newblock Cascade cost volume for high-resolution multi-view stereo and stereo
  matching.
\newblock In {\em Proceedings of the CVPR}, pages 2495--2504, 2020.

\bibitem{hedman2018deep}
Peter Hedman, Julien Philip, True Price, Jan-Michael Frahm, George Drettakis,
  and Gabriel Brostow.
\newblock Deep blending for free-viewpoint image-based rendering.
\newblock {\em ACM Transactions on Graphics}, 37(6):1--15, 2018.

\bibitem{hedman2021baking}
Peter Hedman, Pratul~P Srinivasan, Ben Mildenhall, Jonathan~T Barron, and Paul
  Debevec.
\newblock Baking neural radiance fields for real-time view synthesis.
\newblock {\em arXiv preprint arXiv:2103.14645}, 2021.

\bibitem{huang2018deepmvs}
Po-Han Huang, Kevin Matzen, Johannes Kopf, Narendra Ahuja, and Jia-Bin Huang.
\newblock Deepmvs: Learning multi-view stereopsis.
\newblock In {\em Proceedings of the IEEE Conference on Computer Vision and
  Pattern Recognition}, pages 2821--2830, 2018.

\bibitem{dtu}
Rasmus Jensen, Anders Dahl, George Vogiatzis, Engil Tola, and Henrik Aan{\ae}s.
\newblock Large scale multi-view stereopsis evaluation.
\newblock In {\em 2014 CVPR}, pages 406--413. IEEE, 2014.

\bibitem{jensen2014large}
Rasmus Jensen, Anders Dahl, George Vogiatzis, Engin Tola, and Henrik Aan{\ae}s.
\newblock Large scale multi-view stereopsis evaluation.
\newblock In {\em Proceedings of the IEEE conference on computer vision and
  pattern recognition}, pages 406--413, 2014.

\bibitem{ji2017surfacenet}
Mengqi Ji, Juergen Gall, Haitian Zheng, Yebin Liu, and Lu Fang.
\newblock {SurfaceNet}: An end-to-end {3D} neural network for multiview
  stereopsis.
\newblock In {\em Proc.~ICCV}, 2017.

\bibitem{lsmKarHM2017}
Abhishek Kar, Christian H\"ane, and Jitendra Malik.
\newblock Learning a multi-view stereo machine.
\newblock In {\em NeurIPS}. 2017.

\bibitem{kar2017learning}
Abhishek Kar, Christian H{\"a}ne, and Jitendra Malik.
\newblock Learning a multi-view stereo machine.
\newblock In {\em Proceedings of the 31st International Conference on Neural
  Information Processing Systems}, pages 364--375, 2017.

\bibitem{kopanas2021point}
Georgios Kopanas, Julien Philip, Thomas Leimk{\"u}hler, and George Drettakis.
\newblock Point-based neural rendering with per-view optimization.
\newblock In {\em Computer Graphics Forum}, volume~40, pages 29--43. Wiley
  Online Library, 2021.

\bibitem{lassner2021pulsar}
Christoph Lassner and Michael Zollhofer.
\newblock Pulsar: Efficient sphere-based neural rendering.
\newblock In {\em Proceedings of the IEEE/CVF Conference on Computer Vision and
  Pattern Recognition}, pages 1440--1449, 2021.

\bibitem{li2021neural}
Zhengqi Li, Simon Niklaus, Noah Snavely, and Oliver Wang.
\newblock Neural scene flow fields for space-time view synthesis of dynamic
  scenes.
\newblock In {\em Proceedings of the IEEE/CVF Conference on Computer Vision and
  Pattern Recognition}, pages 6498--6508, 2021.

\bibitem{liu2020neural}
Lingjie Liu, Jiatao Gu, Kyaw~Zaw Lin, Tat-Seng Chua, and Christian Theobalt.
\newblock Neural sparse voxel fields.
\newblock {\em arXiv preprint arXiv:2007.11571}, 2020.

\bibitem{lombardi2019neural}
Stephen Lombardi, Tomas Simon, Jason Saragih, Gabriel Schwartz, Andreas
  Lehrmann, and Yaser Sheikh.
\newblock Neural volumes: Learning dynamic renderable volumes from images.
\newblock {\em arXiv preprint arXiv:1906.07751}, 2019.

\bibitem{luo2019p}
Keyang Luo, Tao Guan, Lili Ju, Haipeng Huang, and Yawei Luo.
\newblock P-mvsnet: Learning patch-wise matching confidence aggregation for
  multi-view stereo.
\newblock In {\em Proceedings of the ICCV}, pages 10452--10461, 2019.

\bibitem{martin2021nerf}
Ricardo Martin-Brualla, Noha Radwan, Mehdi~SM Sajjadi, Jonathan~T Barron,
  Alexey Dosovitskiy, and Daniel Duckworth.
\newblock Nerf in the wild: Neural radiance fields for unconstrained photo
  collections.
\newblock In {\em Proceedings of the IEEE/CVF Conference on Computer Vision and
  Pattern Recognition}, pages 7210--7219, 2021.

\bibitem{mildenhall2020nerf}
Ben Mildenhall, Pratul~P Srinivasan, Matthew Tancik, Jonathan~T Barron, Ravi
  Ramamoorthi, and Ren Ng.
\newblock Nerf: Representing scenes as neural radiance fields for view
  synthesis.
\newblock In {\em European conference on computer vision}, pages 405--421.
  Springer, 2020.

\bibitem{newcombe2011kinectfusion}
Richard~A Newcombe, Shahram Izadi, Otmar Hilliges, David Molyneaux, David Kim,
  Andrew~J Davison, Pushmeet Kohi, Jamie Shotton, Steve Hodges, and Andrew
  Fitzgibbon.
\newblock Kinectfusion: Real-time dense surface mapping and tracking.
\newblock In {\em 2011 10th IEEE international symposium on mixed and augmented
  reality}, pages 127--136. IEEE, 2011.

\bibitem{niessner2013real}
Matthias Nie{\ss}ner, Michael Zollh{\"o}fer, Shahram Izadi, and Marc
  Stamminger.
\newblock Real-time 3d reconstruction at scale using voxel hashing.
\newblock {\em ACM Transactions on Graphics (ToG)}, 32(6):1--11, 2013.

\bibitem{park2021nerfies}
Keunhong Park, Utkarsh Sinha, Jonathan~T Barron, Sofien Bouaziz, Dan~B Goldman,
  Steven~M Seitz, and Ricardo Martin-Brualla.
\newblock Nerfies: Deformable neural radiance fields.
\newblock In {\em Proceedings of the IEEE/CVF International Conference on
  Computer Vision}, pages 5865--5874, 2021.

\bibitem{park2021hypernerf}
Keunhong Park, Utkarsh Sinha, Peter Hedman, Jonathan~T Barron, Sofien Bouaziz,
  Dan~B Goldman, Ricardo Martin-Brualla, and Steven~M Seitz.
\newblock Hypernerf: A higher-dimensional representation for topologically
  varying neural radiance fields.
\newblock {\em arXiv preprint arXiv:2106.13228}, 2021.

\bibitem{pizzoli2014remode}
Matia Pizzoli, Christian Forster, and Davide Scaramuzza.
\newblock Remode: Probabilistic, monocular dense reconstruction in real time.
\newblock In {\em 2014 IEEE International Conference on Robotics and Automation
  (ICRA)}, pages 2609--2616. IEEE, 2014.

\bibitem{scannedobjects}
Google Research.
\newblock Google scanned objects.

\bibitem{ruckert2021adop}
Darius R{\"u}ckert, Linus Franke, and Marc Stamminger.
\newblock Adop: Approximate differentiable one-pixel point rendering.
\newblock {\em arXiv preprint arXiv:2110.06635}, 2021.

\bibitem{russakovsky2015imagenet}
Olga Russakovsky, Jia Deng, Hao Su, Jonathan Krause, Sanjeev Satheesh, Sean Ma,
  Zhiheng Huang, Andrej Karpathy, Aditya Khosla, Michael Bernstein, et~al.
\newblock Imagenet large scale visual recognition challenge.
\newblock {\em International journal of computer vision}, 115(3):211--252,
  2015.

\bibitem{schoenberger2016mvs}
Johannes~Lutz Sch\"{o}nberger, Enliang Zheng, Marc Pollefeys, and Jan-Michael
  Frahm.
\newblock {Pixelwise View Selection for Unstructured Multi-View Stereo}.
\newblock In {\em European Conference on Computer Vision (ECCV)}, 2016.

\bibitem{seitz2006comparison}
Steven~M Seitz, Brian Curless, James Diebel, Daniel Scharstein, and Richard
  Szeliski.
\newblock A comparison and evaluation of multi-view stereo reconstruction
  algorithms.
\newblock In {\em 2006 IEEE computer society conference on CVPR}, volume~1,
  pages 519--528. IEEE, 2006.

\bibitem{nerv2021}
Pratul~P. Srinivasan, Boyang Deng, Xiuming Zhang, Matthew Tancik, Ben
  Mildenhall, and Jonathan~T. Barron.
\newblock Nerv: Neural reflectance and visibility fields for relighting and
  view synthesis.
\newblock In {\em CVPR}, 2021.

\bibitem{sun2021neucon}
Jiaming Sun, Yiming Xie, Linghao Chen, Xiaowei Zhou, and Hujun Bao.
\newblock {NeuralRecon}: Real-time coherent {3D} reconstruction from monocular
  video.
\newblock {\em CVPR}, 2021.

\bibitem{tan2019mnasnet}
Mingxing Tan, Bo Chen, Ruoming Pang, Vijay Vasudevan, Mark Sandler, Andrew
  Howard, and Quoc~V Le.
\newblock Mnasnet: Platform-aware neural architecture search for mobile.
\newblock In {\em Proceedings of the IEEE/CVF Conference on Computer Vision and
  Pattern Recognition}, pages 2820--2828, 2019.

\bibitem{tang2020searching}
Haotian Tang, Zhijian Liu, Shengyu Zhao, Yujun Lin, Ji Lin, Hanrui Wang, and
  Song Han.
\newblock {Searching Efficient 3D Architectures with Sparse Point-Voxel
  Convolution}.
\newblock In {\em European Conference on Computer Vision (ECCV)}, 2020.

\bibitem{ibrnet}
Qianqian Wang, Zhicheng Wang, Kyle Genova, Pratul Srinivasan, Howard Zhou,
  Jonathan~T. Barron, Ricardo Martin-Brualla, Noah Snavely, and Thomas
  Funkhouser.
\newblock Ibrnet: Learning multi-view image-based rendering.
\newblock In {\em CVPR}, 2021.

\bibitem{weder2020routedfusion}
Silvan Weder, Johannes Schonberger, Marc Pollefeys, and Martin~R Oswald.
\newblock Routedfusion: Learning real-time depth map fusion.
\newblock In {\em Proceedings of the IEEE/CVF Conference on Computer Vision and
  Pattern Recognition}, pages 4887--4897, 2020.

\bibitem{wei2021nerfingmvs}
Yi Wei, Shaohui Liu, Yongming Rao, Wang Zhao, Jiwen Lu, and Jie Zhou.
\newblock Nerfingmvs: Guided optimization of neural radiance fields for indoor
  multi-view stereo.
\newblock In {\em ICCV}, 2021.

\bibitem{xiang2021neutex}
Fanbo Xiang, Zexiang Xu, Milos Hasan, Yannick Hold-Geoffroy, Kalyan Sunkavalli,
  and Hao Su.
\newblock Neutex: Neural texture mapping for volumetric neural rendering.
\newblock In {\em Proceedings of the IEEE/CVF Conference on Computer Vision and
  Pattern Recognition}, pages 7119--7128, 2021.

\bibitem{yao2018mvsnet}
Yao Yao, Zixin Luo, Shiwei Li, Tian Fang, and Long Quan.
\newblock {MVS}net: Depth inference for unstructured multi-view stereo.
\newblock In {\em Proc.~ECCV}, pages 767--783, 2018.

\bibitem{yao2019recurrent}
Yao Yao, Zixin Luo, Shiwei Li, Tianwei Shen, Tian Fang, and Long Quan.
\newblock Recurrent mvsnet for high-resolution multi-view stereo depth
  inference.
\newblock In {\em Proceedings of the CVPR}, pages 5525--5534, 2019.

\bibitem{yu2020pixelnerf}
Alex Yu, Vickie Ye, Matthew Tancik, and Angjoo Kanazawa.
\newblock pixelnerf: Neural radiance fields from one or few images.
\newblock In {\em CVPR}, 2021.

\bibitem{zhang2020nerf++}
Kai Zhang, Gernot Riegler, Noah Snavely, and Vladlen Koltun.
\newblock Nerf++: Analyzing and improving neural radiance fields.
\newblock {\em arXiv preprint arXiv:2010.07492}, 2020.

\bibitem{zhou2014color}
Qian-Yi Zhou and Vladlen Koltun.
\newblock Color map optimization for {3D} reconstruction with consumer depth
  cameras.
\newblock {\em ACM Transactions on Graphics}, 33(4):155, 2014.

\end{thebibliography}
}

\appendix
\newpage

\begin{strip}\centering
    \includegraphics[width=\textwidth]{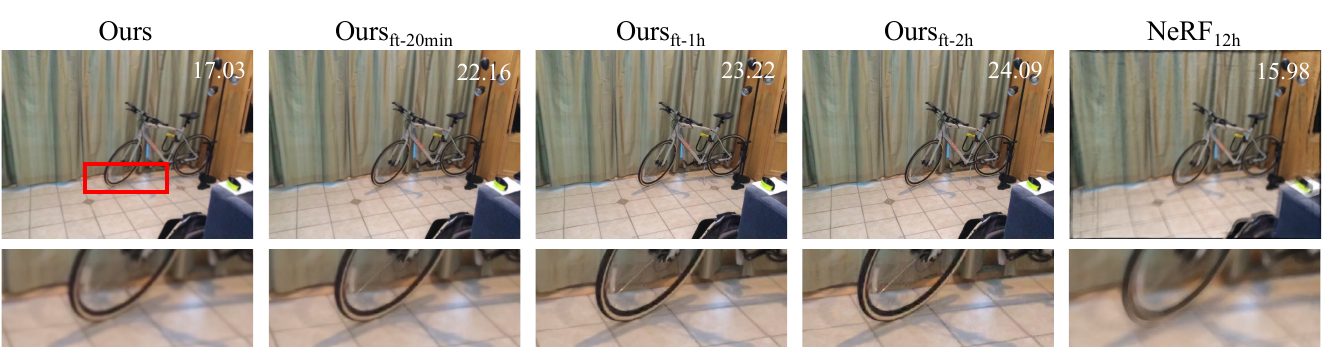}
    \captionof{figure}{Effect of fine-tuning. We show the results with our model with different fine-tuning duration. The first column is our results without fine-tuning. Note that our direct inference result have outperformed NeRF and the results from fine-tuning contain significantly more details. PSNRs are shown at top right of each image.}
    \label{fig:finetune}
\end{strip}

\section{Additional Details}
\label{sec:smodel}

We follow~\cite{sun2021neucon} to use a modified MnasNet~\cite{tan2019mnasnet} pretrained from ImageNet~\cite{russakovsky2015imagenet} as the 2D encoder. The output feature channel number from the 2D encoder is 64. The direction encoder $G$ is a 5 layer MLP with 16-channel output. The SparseConvNet $J$ has 5 SparseConv layers implemented with torchsparse~\cite{tang2020searching}. Each of $M_z, M_r, M_t$ has 3 SparseConv layers. All networks use ReLU as activation layers. All features in the feature volumes (global $\V^g_t$ or local $\V_t$) have 16 channels. We apply positional encoding to the volume feature with maximum frequency $L=5$ before feeding them into the volume renderer. The initial voxel size is set manually for different datasets, specifically $40mm$ for large-scale datasets \eg ScanNet~\cite{dai2017scannet}, and $4mm$ for object-centric datasets \eg NeRF Synthetic~\cite{mildenhall2020nerf} and Google Scanned Objects~\cite{scannedobjects}. When unprojecting 2D features into local feature volumes, we build view frustums with max depth $d_{\max} = 3m$. The pruning threshold $\gamma$ is set to $0.6$ for all experiments. Nearest-neighbor interpolation is used for all coarse-to-fine fine-tuning experiments. The code will be released after the review period.

\section{Additional Results}
\label{sec:sresults}

\begin{figure}\centering
    \includegraphics[width=0.47\textwidth]{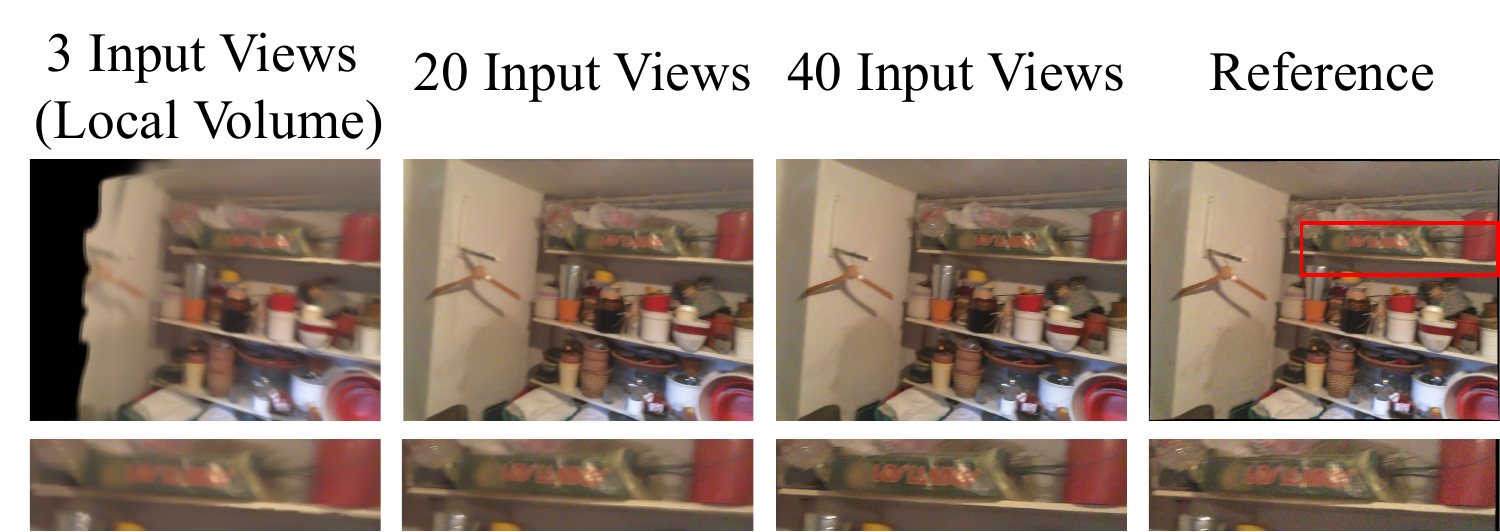}
    \captionof{figure}{Effect of input view numbers. We show the results when sample different numbers of neighboring views to build the feature volume. All results are from the pretrained model without any further fine-tuning. 
    }
    \label{fig:nviews}\vspace{-0.6em}
\end{figure}

\boldstart{Effect of input view numbers.} Figure~\ref{fig:nviews} shows the comparison when sample different numbers of neighboring views to build the feature volume. Our learned global feature fusion module is able to effectively fuse information from different views. Note that, when using more input views, our method could produce sharper details in the rendered images, which are very close to the reference. 

\boldstartspace{Effect of fine-tuning.} Figure~\ref{fig:finetune} shows the visual quality of our method when fine-tuning for different time period. Note that our direct inference result have outperformed NeRF, and the results from fine-tuning contain significantly more details. PSNRs are shown at top right of each image.

\boldstartspace{Geometry reconstruction.} Following~\cite{mildenhall2020nerf,chen2021mvsnerf}, we evaluate our geometry reconstruction quality on the DTU dataset~\cite{jensen2014large} by comparing depth reconstruction results generated from the volume density by a weighted sum of the depth values of the sampled points on marched rays. We use 16 input views and compare the depth quality on both input views and novel views with 2 common metrics. The results are shown in Table~\ref{tab:geo}. Thanks to the explicit geometry modeling in our pipeline, our approach achieves significantly more accurate geometry than other neural rendering methods.

\begin{table}[t]
	\centering
	\begin{tabular}{lcc}
		\toprule
		Method & Abs Err$\downarrow$ & Acc (8mm)$\uparrow$ \\
		\midrule
		PixelNeRF~\cite{yu2020pixelnerf}   &  $0.205/0.211$ & $0.096/0.089$ \\
		IBRNet~\cite{ibrnet}    &  $ 1.123 / 1.324$ & $ 0.000 /0.000$\\
		Ours       &  $\textbf{0.034}/\textbf{0.036}$ & $\textbf{0.722}/\textbf{0.709}$\\
		\bottomrule
\end{tabular}
\rule{0pt}{1.0pt}
\caption{\textbf{Geometry reconstruction.} We evaluate depth reconstruction on the DTU testing set and compare with other two neural rendering methods PixelNeRF~\cite{yu2020pixelnerf} and IBRNet~\cite{ibrnet}. Our method significantly outperforms other neural rendering methods and achieves high depth accuracy. The two numbers of each item refers to the depth at \textit{input} / \textit{novel} views. }
\label{tab:geo}
\end{table}

\end{document}